# Particle Swarm Optimization of Information-Content Weighting of Symbolic Aggregate Approximation

Muhammad Marwan Muhammad Fuad

Department of Electronics and Telecommunications
Norwegian University of Science and Technology (NTNU)
NO-7491 Trondheim, Norway
marwan.fuad@iet.ntnu.no

**Abstract:** Bio-inspired optimization algorithms have been gaining more popularity recently. One of the most important of these algorithms is particle swarm optimization (PSO). PSO is based on the collective intelligence of a swam of particles. Each particle explores a part of the search space looking for the optimal position and adjusts its position according to two factors; the first is its own experience and the second is the collective experience of the whole swarm. PSO has been successfully used to solve many optimization problems. In this work we use PSO to improve the performance of a well-known representation method of time series data which is the symbolic aggregate approximation (SAX). As with other time series representation methods, SAX results in loss of information when applied to represent time series. In this paper we use PSO to propose a new minimum distance WMD for SAX to remedy this problem. Unlike the original minimum distance, the new distance sets different weights to different segments of the time series according to their information content. This weighted minimum distance enhances the performance of SAX as we show through experiments using different time series datasets.

## 1 Introduction

A *time series* is a sequence of real numbers over a period of time. Each of these numbers represents the value of the observed phenomenon at a certain time point. Time series data have been used in many applications such as science, medicine, and engineering. Time series data mining handles several tasks such as similarity search, classification, clustering, and others.

---

This work was carried out during the tenure of an ERCIM "Alain Bensoussan" Fellowship Programme. This Programme is supported by the Marie-Curie Co-funding of Regional, National and International Programmes (COFUND) of the European Commission.

Time series datasets are usually very large so direct search or *sequential scanning* of these datasets is inefficient. In order to overcome this problem transformations can be applied to reduce the dimensionality of the original time series and to represent them in a space with manageable dimensionality. Such transformations are called *dimensionality reduction techniques*, or *representation methods*. The most widely known methods are *Discrete Fourier Transform* (DFT) [2] and [3], *Discrete Wavelet Transform* (DWT) [5], *Singular Value Decomposition* (SVD) [13], *Adaptive Piecewise Constant Approximation* (APCA) [10], *Piecewise Aggregate Approximation* (PAA) [9] and [30], *Piecewise Linear Approximation* (PLA) [17], and *Chebyshev Polynomials* (CP) [4].

Other methods of time series data mining use multi-resolution approaches. In [16] and [27] a method of multi resolution representation of time series is presented. This symbolic method uses a multi-resolution vector quantized approximation of the time series together with a multi-resolution similarity distance. Using this representation the method keeps both local and global information of the time series data. In [19] and [20] other multi-resolution method are proposed. These methods are based on a fast-and-dirty filtering scheme that iteratively reduces the search space using several resolution levels. The technique presented in [22] couples and fast-and-dirty filter with a multi-resolution representation of the time series.

Indexing time series data usually includes establishing a lower bounding distance on time series in the transformed space to guarantee that the representation method will not cause false dismissals. This is achieved by defining a distance, on the transformed space, that underestimates the distance in the original space. This condition is known as the *lower-bounding lemma.* [2].

Among representation methods of time series data, symbolic representation of time series has several advantages which interested researchers in this field of computer science. One of its main advantages is that symbolic representation permits researchers to benefit from the ample symbolic algorithms known in the text-retrieval and bioinformatics communities [14].

There have been many suggestions to represent time series symbolically. But in general, most of these symbolic representation methods suffered from two main inconveniences [15]; the first is that the dimensionality of the symbolic representation method is the same as that of the original space, so there is no virtual dimensionality reduction. The second drawback is that although distance measures have been defined on the reduced symbolic spaces, these distance measures are poorly correlated with the original distance measures defined on the original spaces.

One of the most widely-known symbolic representation methods of time series is SAX. SAX uses pre-computed distances obtained from lookup tables. This makes SAX fast to compute.

In this work we show how the performance of SAX can be improved by substituting the original similarity measure used with SAX by a new one which assigns different weights to different segments of the time series according to their information content. These weights are set using the particle swarm optimization; a widely used population-based optimization method which has been successful in solving many optimization problems. We show through experiments conducted on different time series dataset how the new similarity measure can give better results than the original one.

The work presented in this paper is organized as follows: in Section 2 we present related background. In Section 3 we introduce the new scheme and we evaluate its performance in Section 4. In Section 5 we give concluding remarks.

## 2  Background

### 2.1 The Symbolic Aggregate Approximation (SAX)

The *Symbolic Aggregate approXimation* method (SAX) [14] is one of the most important symbolic representation methods of time series. The main advantage of SAX is that the similarity measure it uses, called MINDIST, uses statistical lookup tables, which makes it is easy to compute with an overall complexity of $O(N)$.

SAX is based on an assumption that normalized time series have Gaussian distribution, so by determining the breakpoints that correspond to a particular alphabet size, one can obtain equal-sized areas under the Gaussian curve. SAX is applied as follows:

1-The time series are normalized.
2-The dimensionality of the time series is reduced using PAA [9], [30]
3-The PAA representation of the time series is discretized by determining the number and location of the breakpoints (The number of the breakpoints is chosen by the user). Their locations are determined, as mentioned above, using Gaussian lookup tables. The interval between two successive breakpoints is assigned to a symbol of the alphabet, and each segment of PAA that lies within that interval is discretized by that symbol.

**Table 1.** The lookup table of $MINDIST$ for alphabet size = 3.

|   | a | b | c |
|---|---|---|---|
| a | 0 | 0 | 0.86 |
| b | 0 | 0 | 0 |
| c | 0.86 | 0 | 0 |

The last step of SAX is using the following similarity measure:

$$MINDIST(\hat{S}, \hat{R}) \equiv \sqrt{\frac{n}{N}} \sqrt{\sum_{i=1}^{N} (dist(\hat{s}_i, \hat{r}_i))^2} \qquad (1)$$

Where $n$ is the length of the original time series, $N$ is the length of the strings (the number of the segments), $\hat{S}$ and $\hat{R}$ are the symbolic representations of the two time series $S$ and $R$, respectively, and where the function $dist(\ )$ is implemented by using the appropriate lookup table. For instance, the lookup table of *MINDIST* for an alphabet size of 3 is the one shown in Table 1.

We also need to mention that the similarity measure used in PAA is:

$$d(S,R) = \sqrt{\frac{n}{N}} \sqrt{\sum_{i=1}^{N}(\overline{s_i} - \overline{r_i})^2} \qquad (2)$$

Where $n$ is the length of the time series, $N$ is the number of segments.

It is proven in [9], [30] that the above similarity distance is a lower bound of the Euclidean distance applied in the original space of time series. This means that *MINDIST* is also a lower bound of the Euclidean distance, because it is a lower bound of the similarity measure used in PAA. This guarantees no false dismissals.

There are other versions and extensions of SAX [11], [25], [26], [28]. These versions use it for other applications or apply it to index massive datasets, or compute *MINDIST* differently [18]. However, the version of SAX that we presented earlier, which is called classic SAX, is the basis of all these versions and extensions and it is actually the most widely-known one.

## 2.2 Information Content

Quantifying the content of information a vector carries was first introduced by Shannon in [24]. This is measured by what is known as *entropy* and is defined for a discrete probabilistic system by:

$$H = -\sum_{i} p_i \log p_i \qquad (3)$$

where the base of the logarithm is 2.

This concept has many applications in cryptography, data transmission, natural language processing, data compression, and others.

In time series mining, the concept of information content was implicitly or explicitly present in different representation methods of time series data. DFT [2], [3] and DWT [5], for instance are based on the fact that the first coefficients are the most meaningful ones; i.e. they contain most of the information in the time series, so the other coefficients can be truncated without much loss of information. APCA [10] segments the time series into segments of varying lengths such that their individual reconstruction errors are minimal. The intuition behind this idea is that different regions of the time series contain different amounts of information. So while regions of high activity contain high fluctuations, other regions of low activity show a flat behavior, so a representation method with high fidelity should reflect this difference in behavior.

## 3 Particle Swarm Optimization of Information-Content Weighting of Symbolic Aggregate Approximation (PSOWSAX)

In time series mining the distance that is widely used to compute the similarity between the two time series $S = \{s_1, s_2, ..., s_n\}$ and $R = \{r_1, r_2, ..., r_n\}$ is the Euclidean distance which is defined as follows:

$$L_2(S, R) = \sqrt[2]{\sum_{i=1}^{n} |s_i - r_i|^2} \qquad (4)$$

One of the variations of the Euclidean distance, which is much related to the topic of this paper, is the *Weighted Euclidean Distance*. This distance is defined as:

$$d(S, R, W) = \sqrt{\sum_{i=1}^{n} w_i (s_i - r_i)^2} \qquad (5)$$

where $W$ is the weight vector.

Fig. 1 shows the Euclidean distance and the weighted Euclidean distance between two time series.

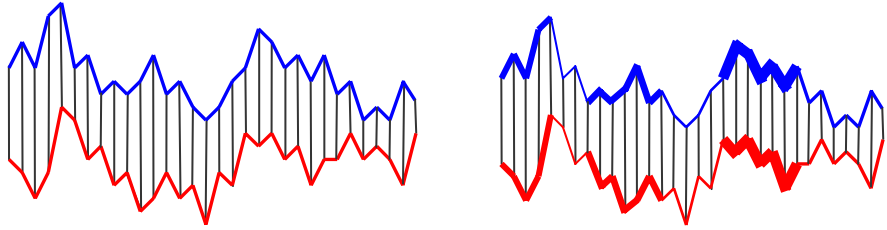

**Fig. 1.** The Euclidean distance (left) and the weighted Euclidean distance (right)

The intuition behind the weighted Euclidean distance is, again, that some parts of the time series may have more importance than other parts so the distance should reflect these regions differently compared with regions that contain less information.

Although weighing different regions differently seems to be a good solution to distinguish parts with high information content from others with less information, the question remains on how to set the weights. In [29] the authors proposed setting the weights using relevance feedback provided by the user. We can easily see that this solution is highly subjective and also inefficient.

It is important to mention that the weighted distance defined in (5) is applied to the raw time series and not to the reduced, lower-dimensional time series used in representation methods.

In this paper we present a modification of SAX based on the concept of information content.

In Section 2.1, we presented *MINDIST* ; the similarity measure used with SAX. From relation (1) we can derive the following similarity measure which we call the *Weighted Minimum Distance* (*WMD*):

$$WMD(\hat{S},\hat{R}) = \sqrt{\frac{n}{N}} \sqrt{\sum_{i=1}^{N} w_i (dist(\hat{s}_i, \hat{r}_i))^2} ; \qquad w_i \in [0,1] \qquad (6)$$

Notice that if we set $w_i = 1$, $\forall i$ in (6) we obtain *MINDIST* defined in relation (1), so *MINDIST* is in fact a special case of *WMD*. Notice also, which is very important, that from (1) and (6) we can easily see that:

$$WMD(\hat{S},\hat{R}) \leq MINDIST(\hat{S},\hat{R}) \qquad (7)$$

Since *MINDIST* is a lower bound of the Euclidean distance in the original space this implies that *WMD* is also a lower bound of the Euclidean distance, so our proposed distance guarantees no false dismissals. .

The weights in (6) will be set for the whole time series in the dataset using the particle swarm optimization.

**Particle Swarm Optimization:** *Particle Swarm Optimization* (PSO) is a member of a family of naturally-inspired optimization algorithms called *Swarm Intelligence* which are population-based optimization algorithms. PSO was inspired by the social behavior of some animals, such as bird flocking or fish schooling [8]. [23] proposed a model that simulates a swarm. In this model individuals, also called *agents* or *particles*, follow these rules (Fig. 2):

Separation: Each particle avoids getting too close to its neighbors.
Alignment: Each particle steers towards the general heading of its neighbors.
Cohesion:  Each particle moves towards the average position of its neighbors.

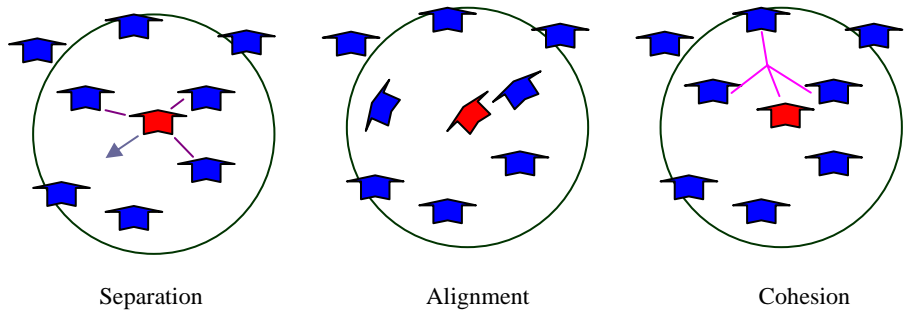

Separation　　　　　　　　Alignment　　　　　　　　Cohesion

**Fig. 2.** Simulating swarm's behavior

The above model was elaborated by adding another rule which is *obstacle avoidance* which uses *steer-to-avoid* concept.

As is the case with many other optimization algorithms, there are quite a large number of variations of PSO. In the following we present a standard PSO [7]. PSO starts by initializing a swarm of *sSize* particles at random positions $\vec{X}_i^0$ and velocities $\vec{V}_i^0$ where $i \in \{1,...,sSize\}$.

In the next step each position is evaluated, and for each iteration, using a *fitness function,* also called *objective function* or *cost function.*

The positions $\vec{X}_i^{k+1}$ and velocities $\vec{V}_i^{k+1}$ are updated at time step ($k+1$) according to the following formulae:

$$\vec{V}_i^{k+1} = \omega.\vec{V}_i^k + \varphi_G\left(\vec{G}^k - \vec{X}_i^k\right) + \varphi_L\left(\vec{L}_i^k - \vec{X}_i^k\right) \qquad (8)$$

$$\vec{X}_i^{k+1} = \vec{X}_i^k + \vec{V}_i^k \qquad (9)$$

where $\varphi_G = r_G.a_G$, $\varphi_L = r_L.a_L$, $r_G, r_L \to U(0,1)$, $\omega, a_L, a_G \in \mathbf{R}$, $\vec{L}_i^k$ is the best position found by particle $i$, $\vec{G}^k$ is the global best position found by the whole swarm, $\omega$ is called the *inertia*, $a_L$ is called the *local acceleration*, and $a_G$ is called the *global acceleration*. These last three parameters are control parameters which are chosen by the algorithm designer.

---

**Algorithm1** Particle Swarm Optimization (PSO)

**Require** Number of parameters (*nPar*), swarm size
(*sSize*), number of iterations (*nItr*), local acceleration
($a_L$), global acceleration ($a_G$), inertia $\omega$.

```
1: Initialize X⃗ᵢ⁰, V⃗ᵢ⁰
2: for itr=1 to nItr do
3:    for all particles Pᵢ do
4:       r_L, r_G ← rand
5:       Update the particle's velocity:
6:       V⃗ᵢ ← ωV⃗ᵢ + φ_G(G⃗ − X⃗ᵢ) + φ_L(L⃗ᵢ − X⃗ᵢ)
7:       Move the particle to the new position:
8:       X⃗ᵢ ← X⃗ᵢ + V⃗ᵢ
9:       if f(X⃗ᵢ) ≤ f(L⃗ᵢ) then X⃗ᵢ ← L⃗ᵢ
10:      if f(X⃗ᵢ) ≤ f(G⃗ᵢ) then X⃗ᵢ ← G⃗ᵢ
11:   end for
12: end for
```

---

**Fig. 3.** The particle swarm optimization algorithm

The algorithm continues until a stopping criterion terminates it. Fig. 3 presents a pseudo code of PSO.

As in the case with other evolutionary algorithms, PSO should keep a balance between *exploitation,* and *exploration*. Exploration is defined as the act of searching for the purpose of discovery, and exploitation is defined as the act of utilizing something for any purpose [1].

Diversity in PSO comes from two sources [31]; the first is the difference between the particle's current position and that of its best neighbor, and the other is the difference between the particle's current position and its best historical position. Variation, although provides exploration, can only be sustained for a limited number of generations because convergence of the swarm to the best position is necessary to refine the solution (exploitation).

## 4  Experimental Validation

We tested our distance on a time series classification task on the datasets available at [12], which is the same repository on which the original SAX was tested. This repository makes up between 90% and 100% of all publicly available, labeled time series data sets in the world, and it represents the interest of the data mining/database community, and not just one group [6].

We tested our method in a classification task based on the first nearest-neighbor (1-NN) rule using leaving-one-out cross validation. This means that every time series is compared to the other time series in the dataset. If the 1-NN does not belong to the same class, the error counter is incremented by 1.

The purpose of the experiments is to compare PSOWSAX (our new method which uses *WMD* as a similarity measure) with the original method (which uses *MINDIST* as a similarity measure) on the classification task and see which one gives the smallest error. This means that for each value of the alphabet size tested we perform the three steps of SAX presented in Section 2.1 to obtain the symbolic representation of the time series, and then we apply *MINDIST* when we test the original method or we apply *WMD* when we want to test ours, which, as indicated earlier, is based on PSO. The weights $w_i$ in relation (6) are obtained during a training phase. This means for each value of the alphabet size tested, we formulate a PSO-based optimization problem where the fitness function is the classification error of the time series (we opt to minimize the classification error) and the outcome of this optimization problem is the weights $w_i$ that yield this minimum value of the classification error, then we use these values on the corresponding testing datasets to obtain the final classification error. As for *MINDIST*, there is no training phase and it is applied directly to the testing datasets.

For PSOWSAX, the swarm size we used in the experiments is 16. Each particle is a vector of *nPar* components representing potential weights of the segments of the time series.

We used a standard PSO, the local acceleration $a_L$ was set to 2, the global acceleration $a_G$ was set to 2. The dimension of the problem $nPar$ is the dimension of the weight vector, which is the number of segments of the times series in the reduced space; i.e. $nPar = \dfrac{length\ of\ the\ original\ time\ series}{compression\ ratio}$. This dimension differs from one dataset to another. The value of inertia $\omega$ depends on the current iteration according to the formula $\omega = (nItr - itr)/nItr$ where $itr$ is the current iteration. This means that the influence of $\omega$ decreases as the algorithm progresses.

The number of iterations $nItr$ was set to 20. This is a rather small number of iterations and the algorithm could have been left to evolve more. However, the objective of our experiments was to validate our method rather than to show its best performance which could be achieved using more sophisticated PSO algorithms.

We also have to mention that we know from experience that for some datasets the classification error is always high, or always low, for all methods of time series representation, so a threshold of a classification error, related to the dataset tested, could be set as a stopping criterion.

Table 2 summarizes the symbols used in the experiments together with their corresponding values.

**Table 2.** The symbol table together with the corresponding values used in the experiments

| | | |
|---|---|---|
| $sSize$ | Swarm size | 16 |
| $nItr$ | Number of iterations | 20 |
| $a_L$ | Local acceleration | 2 |
| $a_G$ | Global acceleration | 2 |
| $\omega$ | Inertia | $\omega = (nItr - itr^*)/nItr$ |

$itr^*$ : The current iteration.

In Fig. 4. we present some of the results we obtained for alphabet size equal to 3, 10, and 20, respectively. We chose to report these values because the first version of SAX used alphabet size that varied between 3 and 10. Then in a later version the alphabet size varied between 3 and 20. So these values are benchmark values.

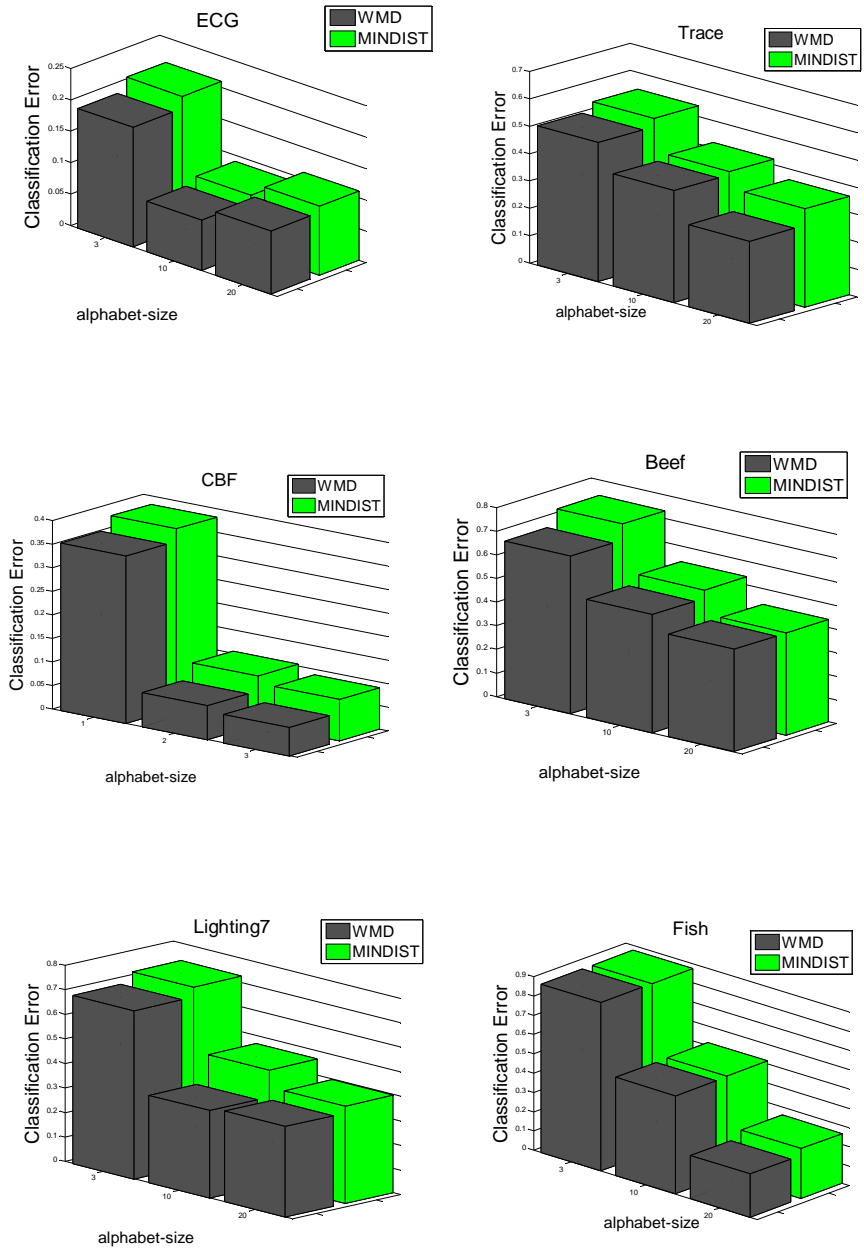

**Fig. 4.** The classification errors obtained by using *WMD* and *MINDIST*

As we can see from Fig. 4, the classification error of *WMD* is smaller than that of *MINDIST* for all values of the alphabet size and for all the datasets, except for dataset (Beef), alphabet size=20 and dataset (Fish), alphabet size=3 where both *WMD* and *MINDIST* gave the same classification error.

Finally, we present in Table 3, for reproducibility purposes, the weights of datasets (CBF) and (ECG) (Because of space restrictions, we present these datasets only). As indicated earlier, these weights are obtained by applying PSOWSAX to the training datasets. The final classification errors of *WMD* (those shown in Fig. 4) are obtained by using those weights, with *WMD*, on the corresponding testing datasets.

**Table 3.** The weights of different segments of the time series obtained by using PSOWSAX

| Dataset | Alphabet Size | $w_i$ |
|---|---|---|
| CBF | 3 | [0.69  0.256  0.921  0.537  0.105  0.383  0.856  0.581  0.915  0.476  0.426  0.899  0.558  0.565  0.021  0.809  0.381  0.394  0.03  0.556  0.616  0.084  0.955  0.519  0.358  0.143  0.221  0.197  0.086  0.485  0.264  0.159] |
| CBF | 10 | [0.722  0.331  0.073  0.662  0.51  0.005  0.107  0.943  0.049  0.869  0.667  0.56  0.396  0.643  0.165  0.573  0.968  0.304  0.492  0.342  0.548  0.372  0.804  0.608  0.874  0.159  0.452  0.514  0.524  0.47  0.488  0.003] |
| CBF | 20 | [0.525  0.408  0.311  0.676  0.154  0.215  0.106  0.89  0.575  0.803  0.411  0.543  0.427  0.648  0.639  0.358  0.719  0.03  0.69  0.63  0.547  0.043  0.553  0.313  0.564  0.740  0.798  0.711  0.847  0.173  0.659  0.447] |
| ECG | 3 | [0.629 0.538 0.186 0.584 0.772 0.106 0.929 0.143 0.639 0.958 0.409 0.567  0.794  0.2  0.704  0.3  0.351  0.982  0.598  0.337  0.6  0.3  0.662  0.395] |
| ECG | 10 | [0.552 0.638 0.967 0.923 0.19 0.027 0.540 0.951 0.966 0.715 0.264 0.857  0.145 0.759 0.901 0.418 0.463 0.358 0.408 0.281 0.918 0.817 0.366 0.0344] |
| ECG | 20 | [0.572  0.612  0.804  0.062  0.268  0.685  0.428  0.546  0.3  0.359  0.679  0.408  0.564  0.034  0.797  0.152  0.683  0.758  0.563  0.091  0.931  0.582  0.044  0.006] |

## 5 Conclusion

In this paper we showed through a new scheme PSOWSAX, based on particle swarm optimization, how the performance of SAX; one of the most important symbolic representation methods of time series data, can be improved by using a new similarity measure *WMD* which assigns different weights to different segments of the time series according to their information content. The information loss caused by time series representation methods can be better recovered by setting different weights to different regions of the time series according to their information content.

The optimization process takes place at indexing time so the new scheme has the same low complexity as that of the original SAX.

We validated the new scheme by conducting classification task experiments on different datasets. The experiments showed that our new scheme gives better results than the original one.

A possible future work will be to associate the work presented in this paper with the work presented in [20]. This can be achieved in two ways; the first is to use the optimization scheme presented in [20] to locate the breakpoints then to use the optimization scheme presented in this work to set different weights to different segments according to their information content. The second way is to use a one-step optimization problem to locate the breakpoints together with the corresponding weights of segments.

## References

1. Agnes, M., Webster's New World College Dictionary, Webster's New World, ISBN 0764571257, May 2004. (2004)
2. Agrawal, R., Faloutsos, C., and Swami, A.: Efficient Similarity Search in Sequence Databases. Proceedings of the 4th Conf. on Foundations of Data Organization and Algorithms. (1993)
3. Agrawal, R., Lin, K. I., Sawhney, H. S. and Shim, K.: Fast Similarity Search in the Presence of Noise, Scaling, and Translation in Time-Series Databases. In Proceedings of the 21st Int'l Conference on Very Large Databases. Zurich, Switzerland, pp. 490-501(1995)
4. Cai, Y., and Ng, R. : Indexing Spatio-temporal Trajectories with Chebyshev Polynomials. In SIGMOD (2004)
5. Chan, K. and Fu, A. W.: Efficient Time Series Matching by Wavelets. In proc. of the 15th IEEE Int'l Conf. on Data Engineering. Sydney, Australia, Mar 23-26. pp 126-133. (1999)
6. Ding, H., Trajcevski, G., Scheuermann, P., Wang, X., and Keogh, E.: Querying and Mining of Time Series Data: Experimental Comparison of Representations and Distance Measures. In Proc of the 34th VLDB (2008)
7. Fernández-Martínez, J.L. and García-Gonzalo, E.: What Makes Particle Swarm Optimization a Very Interesting and Powerful Algorithm? In B.K. Panigrahi, Y. Shi, and M.-H. Lim (Eds.): Handbook of Swarm Intelligence, ALO 8, pp. 37–65 (2011)
8. Haupt, R.L., Haupt, S. E.: Practical Genetic Algorithms with CD-ROM. Wiley-Interscience (2004)
9. Keogh, E,. Chakrabarti, K,. Pazzani, M. and Mehrotra: Dimensionality Reduction for Fast Similarity Search in Large Time Series Databases. J. of Know. and Inform. Sys. (2000)
10. Keogh, E,. Chakrabarti, K,. Pazzani, M., and Mehrotra, S. : Locally Adaptive Dimensionality Reduction for Similarity Search in Large Time Series Databases. SIGMOD pp 151-162 (2001)
11. Keogh, E., Lin, J., and Fu, A.: HOT SAX: Efficiently Finding the Most Unusual Time Series Subsequence. In Proc. of the 5th IEEE International Conference on Data Mining (ICDM 2005), Houston, Texas, Nov 27-30 (2005)
12. Keogh, E., Zhu, Q., Hu, B., Hao. Y., Xi, X., Wei, L. & Ratanamahatana, The UCR Time Series Classification/Clustering Homepage: www.cs.ucr.edu/~eamonn/time_series_data/
C. A. (2011)
13. Korn, F., Jagadish, H., and Faloutsos, C.: Efficiently Supporting Ad Hoc Queries in Large Datasets of Time Sequences. Proceedings of SIGMOD '97, Tucson, AZ, pp 289-300 (1997)
14. Lin, J., Keogh, E., Lonardi, S., Chiu, B. Y.: A Symbolic Representation of Time Series, with Implications for Streaming Algorithms. DMKD 2003: 2-11(2003)
15. Lin, J., Keogh, E., Wei, L., and Lonardi, S.: Experiencing SAX: a Novel Symbolic Representation of Time Series. DMKD Journal (2007)

16. Megalooikonomou, C.: Multiresolution Symbolic Representation of Time Series. In proceedings of the 21st IEEE International Conference on Data Engineering (ICDE). Tokyo, Japan. (2005)
17. Morinaka, Y., Yoshikawa, M. , Amagasa, T., and Uemura, S.: The L-index: An Indexing Structure for Efficient Subsequence Matching in Time Sequence Databases. In Proc. 5th PacificAisa Conf. on Knowledge Discovery and Data Mining, pages 51-60 (2001)
18. Muhammad Fuad, M.M., Marteau, P.F.: Enhancing the Symbolic Aggregate Approximation Method Using Updated Lookup Tables. 14th International Conference on Knowledge-Based and Intelligent Information & Engineering Systems – KES 2010. Cardiff, Wales, UK. September 8-10 (2010)
19. Muhammad Fuad, M.M., Marteau, P.F.: Fast Retrieval of Time Series by Combining a Multiresolution Filter with a Representation Technique. The International Conference on Advanced Data Mining and Applications–ADMA2010, ChongQing, China, 21 November (2010)
20. Muhammad Fuad, M.M. : Genetic Algorithms-Based Symbolic Aggregate Approximation. 14th International Conference on Data Warehousing and Knowledge Discovery - DaWaK 2012 – Vienna, Austria, September 3 – 7 (2012)
21. Muhammad Fuad, M.M., Marteau, P.F.: Multi-resolution Approach to Time Series Retrieval. Fourteenth International Database Engineering and Applications Symposium– IDEAS 2010 , 16-18 August, 2010, Montreal, QC, CANADA (2010)
22. Muhammad Fuad, M.M., Marteau, P.F.: Speeding-up the Similarity Search in Time Series Databases by Coupling Dimensionality Reduction Techniques with a Fast-and-dirty Filter. Fourth IEEE International Conference on Semantic Computing– ICSC 2010, 22-24 September 2010, Carnegie Mellon University, Pittsburgh, PA,  USA (2010)
23. Reynolds, C. W. : Flocks, Herds and Schools: A Distributed Behavioral Model. SIGGRAPH Comput. Graph. 21, 4 (1987)
24. Shannon, C.: A Mathematical Theory of Communication. The Bell Systems Technical Journal 27, 379–423, 623–656 (1948)
25. Shieh, J. and Keogh, E.: iSAX: Disk-Aware Mining and Indexing of Massive Time Series Datasets. Data Mining and Knowledge Discovery (2009)
26. Shieh, J. and Keogh, E.: iSAX: Indexing and Mining Terabyte Sized Time Series. In Proceeding of the 14th ACM SIGKDD International Conference on Knowledge Discovery and Data Mining, Las Vegas, Nevada, USA, August 24 – 27 (2008)
27. Wang, Q., Megalooikonomou, V., and Faloutsos, C.: Time Series Analysis with Multiple Resolutions. Inf. Syst. 35, 1 (2010)
28. Wei, L., Keogh, E., and Xi, X.: SAXually Explict Images: Finding Unusual Shapes. ICDM (2006)
29. Wu, L., Faloutsos, C., Sycara, K., Payne, T.: FALCON: Feedback Adaptive Loop for Content-Based Retrieval VLDB 2000: 297-306 (2000)
30. Yi, B. K., and Faloutsos, C.: Fast Time Sequence Indexing for Arbitrary Lp Norms. Proceedings of the 26th International Conference on Very Large Databases, Cairo, Egypt (2000)
31. Zavala, A.M., Aguirre, A.H., and Diharce, E.V.: Robust PSO-based Constrained Optimization by Perturbing the Particle's Memory. Swarm Intelligence: Focus on ant and particle swarm optimization, Felix T. S. Chan and Manoj Kumar Tiwari,Ed. I-Tech Education and Publishing (2007)